\pdfoutput=1
\documentclass[10pt,twocolumn,letterpaper]{article}
\usepackage[hyphens]{url}
\usepackage{todonotes}
\usepackage{dsfont}
\usepackage{cvpr}
\usepackage{times}
\usepackage{epsfig}
\usepackage{graphicx}
\usepackage{amsmath}
\usepackage{amssymb}
\usepackage[numbers,square,compress,sort]{natbib}
\usepackage{tabularx}
\usepackage{javen}
\usepackage{booktabs}
\usepackage[export]{adjustbox}
\usepackage{epstopdf}

\newcommand*\samethanks[1][\value{footnote}]{\footnotemark[#1]}

\usepackage[pagebackref=true,breaklinks=true,letterpaper=true,colorlinks,bookmarks=false]{hyperref}

\cvprfinalcopy 

\def\cvprPaperID{3362} 

\ifcvprfinal\fi
\begin{document}

\title{ Convolutional Sequence to Sequence Model for Human Dynamics}

\author{Chen Li\thanks{Considered as equal contribution.}\quad\quad Zhen Zhang\samethanks[1] \quad\quad Wee Sun Lee\quad\quad Gim Hee Lee\\
  Department of Computer Science,\\
  National University of Singapore, Singapore\\
  {\tt\small \{lic,zhangz,leews,gimhee.lee\}\@comp.nus.edu.sg}
}

\maketitle

\begin{abstract}
  Human motion modeling is a classic problem in computer vision and graphics. Challenges in modeling human motion include high dimensional prediction as well as extremely 
complicated dynamics.
We present a novel
approach to human motion modeling based on convolutional neural
networks (CNN). The hierarchical structure of CNN makes it capable of
capturing both spatial and temporal correlations effectively.
In our proposed approach, 
a convolutional long-term encoder is used to encode the whole
given motion sequence into a long-term hidden variable, which is used with a
decoder to predict the remainder of the sequence. The decoder itself
also has an encoder-decoder structure, in which the short-term
encoder encodes a shorter sequence to a short-term hidden variable, and
the spatial decoder maps the long and short-term hidden variable to
motion predictions. By using such a model, we are able to capture both
invariant and dynamic information of human motion, which results in
more accurate predictions. 
  Experiments show that our algorithm outperforms the
state-of-the-art methods on the Human3.6M and CMU Motion Capture datasets. Our code is available at the project website\footnote{\url{https://github.com/chaneyddtt/Convolutional-Sequence-to-Sequence-Model-for-Human-Dynamics}}.
\end{abstract}



\begin{figure}[t]
  \centering
  \includegraphics[width=0.5\textwidth]{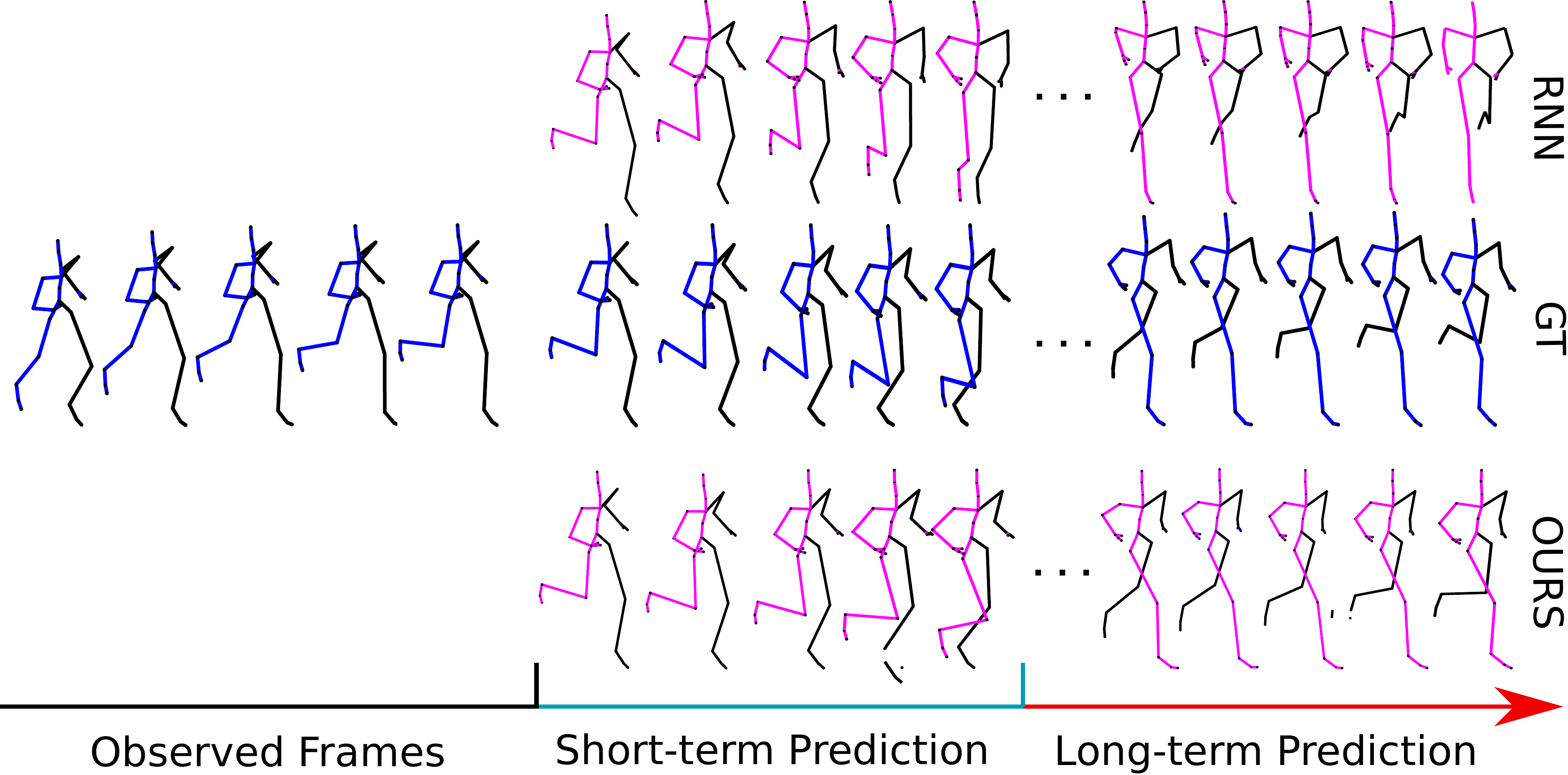}
  \caption{Frames on the left are the observations fed into our network. The middle part is the short-term prediction results for RNN (on the top) and our model (in the bottom). The right part is the long-term prediction results, in which RNN converge to a mean pose, while our model can predict future frames which are similar to the ground truth (in the middle).}
  \label{fig:target}
\end{figure}
\section{Introduction}
\label{sec:introduction}


Understanding
human motion is extremely  
important 
for various applications in
computer vision and robotics, particularly for applications
that require interaction with humans. For example, an unmanned vehicle 
must have the ability to predict human motion in order to avoid
potential collision in a crowded street. Besides, applications such as
sports analysis and medical diagnosis may also benefit 
from human motion modeling.

The biomechanical dynamics of human motion is extremely
complicated. Although several analytic models have been proposed, they
are limited to few 
simple actions such as standing and walking \citep{tozeren1999human}. For more complicated actions, data-driven
methods are required to attain acceptable accuracy
\citep{Martinez_2017_CVPR,tozeren1999human,Butepage_2017_CVPR}. 
In this paper, we focus on the human motion prediction task, using
learning based methods from motion capture data.


Recently, along with the success of deep learning in various areas of
computer vision and machine learning, deep recurrent neural network based models 
have been introduced in human motion prediction
\citep{ghosh2017learning, Martinez_2017_CVPR, jain2016structural,
  fragkiadaki2015recurrent}. 
In earlier works
\citep{fragkiadaki2015recurrent,jain2016structural}, it is often
observed that there is a significant discontinuity between the first predicted
frame of
motion 
and the last frame of the true observed motion. Martinez \etal \citep{Martinez_2017_CVPR} solved the
problem by adding a residual unit in the recurrent network. However,
their residual unit based model often converges to an undesired mean pose in the
long-term predictions, i.e., the predictor
gives static predictions similar to the mean of the ground truth of future sequences (see Figure~\ref{fig:target}). 
We believe that the mean pose problem
is caused by the fact that it is difficult for recurrent models to learn to keep
track of long-term information; the mean pose becomes a good prediction of the future pose when the model loses track of information from the distant past.
For a chain-structured RNN model,  it takes
$n$ steps for two elements that are $n$ time steps apart to interact with each
other; this may make it difficult for an RNN to learn a structure that is able to exploit long-term correlations \citep{pmlr-v70-gehring17a}.

The current state-of-the-art for human motion modeling \citep{Martinez_2017_CVPR} is based on the
sequence-to-sequence model, which is first proposed for
machine translation \citep{sutskever2014sequence}. The
sequence-to-sequence model consists of an encoder and a decoder, in which
the encoder maps a given seed
sequence to a hidden variable, and the decoder maps the hidden
variable to the target sequence.  A major difference between
human motion prediction and other sequence-to-sequence tasks is that
human motion is a highly constrained system 
by environment 
properties, human body properties and Newton's Laws. As a result, the
encoder needs to learn these constraints from a relative long seed
sequence. However, RNN may not be able to learn these constraints accurately, and the accumulation of these errors in decoder may
result in larger error in long-term prediction.

Furthermore, the human body is often not static stable during motion,
and our central neural system must make multiple parts of our body
coordinate with each other to stabilize the motion under gravity and
other loads \citep{winter1995human}. Thus joints from different limbs
have both temporal and spatial 
correlations. A typical
example is human walking. During walking, most people 
tend to 
move left arm forward while moving right leg forward. However,
RNN based methods have difficulties learning to capture this kind of spatial correlations well, and
thus may generate some unrealistic predictions. 
Despite these limitations, the RNN based
 method \citep{Martinez_2017_CVPR} is 
 considered to be the current
 state-of-the-art as its performance is superior to other human motion
 prediction methods in terms of accuracy.


 In this paper, 
 we build a convolutional sequence-to-sequence model for
 the human motion prediction problem. Unlike previous chain-structured
 RNN models, the hierarchical structure of convolutional neural networks
 allows it to naturally 
 model and learn both spatial dependencies as well as long-term temporal 
 dependencies \citep{pmlr-v70-gehring17a}. 
 We evaluate the proposed method on the Human3.6M and the CMU Motion Capture
 datasets. Experimental results show that our method 
can better avoid the long-term mean pose problem, and give more realistic predictions. The quantitative results also show that our algorithm outperforms state-of-the-art methods in terms of accuracy.











\section{Related Works}
\label{sec:related-works}
The main task of this work is human motion prediction via
convolutional models, while previous works\citep{Butepage_2017_CVPR,Martinez_2017_CVPR,fragkiadaki2015recurrent} mainly focus on RNN based
models. We briefly review the literature as follows.

\paragraph{Modeling of human motion} Data driven methods in human
motion modeling face a series of difficulties including
high-dimensionality, complicated non-linear dynamics and the
uncertainty of human movement. Previously, Hidden Markov
Model \citep{brand2000style}, linear dynamics system
\citep{pavlovic2001learning}, Gaussian Process latent variable models
\citep{wang2008gaussian} \etc have been applied to model human
motion. Due to limited computational resource, all these models have
some trade-off between model capacity and inference
complexity. Conditional Restricted Boltzmann Machine (CRBM) based
method 
has also been applied to
human motion modeling \citep{taylor2007modeling}. However, CRBM
requires a more complicated training process and it also requires 
sampling
for approximate inference. 


\paragraph{RNN based human motion prediction} Due to the success of
recurrent models in sequence-to-sequence learning, a series of
recurrent neural network based methods are proposed for the human
motion prediction
task 
\citep{fragkiadaki2015recurrent,Martinez_2017_CVPR,jain2016structural}.
Most of these works have a recurrent network based
encoder-decoder structure, where the encoder accepts a given motion
frames or sequence and propagates 
an encoded hidden variable to the 
decoder, 
  which then generates the future motion frame or series. 
The main differences in
these works lie in their different encoder and decoder structures. 
For example, in the
Encoder-Recurrent-Decoder (ERD) model
\citep{fragkiadaki2015recurrent},  additional non-recurrent spatial
encoder and decoder are 
are added to its recurrent part, which captures the temporal dependencies
\citep{sutskever2014sequence}. In Structural-RNN
\citep{jain2016structural}, several RNNs are stacked together according to a
hand-crafted spatial-temporal graph. Martinez \etal \cite{Martinez_2017_CVPR} proposed a residual
based model, which predicts the gradient of human motion rather than
human motion directly, and used a standard sequence-to
-sequence learning model with 
(GRU) \citep{cho2014properties} cell as the encoder and decoder. In these RNN
based models, fully-connected layers are used to learn a
representation of human action, and the recurrent middle layers 
are
applied to model the temporal dynamics. 
In contrast to previous RNN based models, we use a convolutional model
to learn the spatial and the temporal dynamics at the same time and we show that this outperforms the state-of-the-art methods in human motion
prediction.

\paragraph{Convolutional sequence-to-sequence model}
The task of a sequence-to-sequence model is to generate a target sequence from a
given seed sequence. Most sequence-to-sequence models consist of two
parts, an encoder which encodes the seed sequence into a hidden variable
and a decoder which generates the target sequence from the hidden
variable. 
Although RNNs
seem to be a natural choice for sequential data,  
convolutional models have also been adopted to sequence-to-sequence
tasks such as machine translation. Kalchbrenner and Blunsom \citep{kalchbrenner2013recurrent}
proposed the Recurrent Continuous 
Translation Model (RCTM) which uses a convolutional model as encoder to
generate hidden variables 
and a RNN as decoder to generate target sequences, while later
methods \citep{bradbury2016quasi,
  pmlr-v70-gehring17a,Kalchbrenner2016NeuralMT} are fully
convolutional sequence-to-sequence model. However, unlike machine
translation, where only temporal
correlations exist, there exist complicated spatial-temporal dynamics in human motion. Thus
we design a convolutional sequence-to-sequence model that is suitable for complicated spatial-temporal dynamics.  

\section{Network Architecture}
\label{sec:methology}

\begin{figure*}[t]
  \centering
  \includegraphics[width=0.9\textwidth]{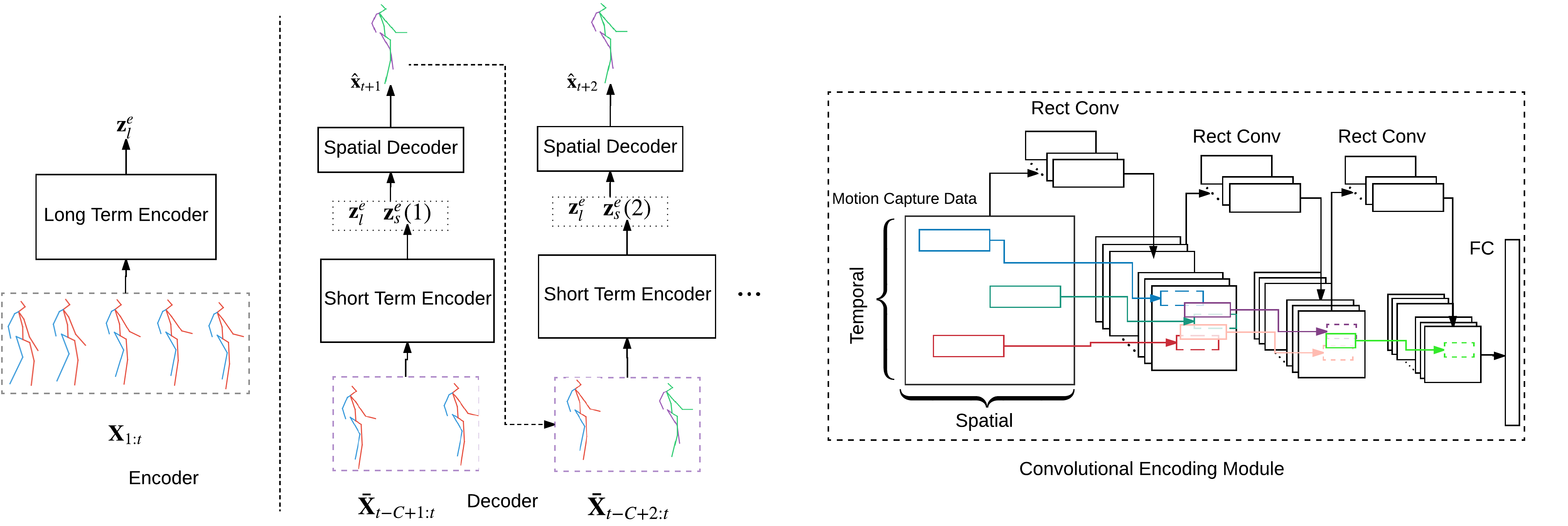}
  \caption{An illustration of our network architecture.}
  \label{fig:arch}\vspace{-1em}
\end{figure*}


We adapt a multi-layer convolutional architecture,
which has the advantage of expressing input
sequences hierarchically. In particular, when we apply convolution
to
the
input skeleton sequences, lower layers will capture 
dependencies between nearby frames and higher layers will capture
dependencies between distant frames. Unlike the chain-structured RNN, the hierarchical
structure of a multi-layer convolutional architecture is designed to 
capture
long-term dependencies.
%
Figure~\ref{fig:arch} shows an illustration of the architecture of our network, where the convolutional encoding module (CEM) plays the central role. We use the CEM module as long and short-term encoder. The long-term encoder is used to memorize a given motion sequence as the long-term hidden variable $\zb_l^e$, and the short-term encoder is used to map a shorter sequence to the short-term hidden variable $\zb_s^e$. Finally the hidden variables $\zb_l^e$ and $\zb_s^e$ are concatenated together and propagated to the decoder to give a prediction of next frame. The short-term encoder and decoder are applied recursively to produce the whole predicted sequence. We use convolutional layers with stride two in the CEM, thus two elements with distance $n$ are able to interact with each other in $\Ocal(\log n)$ operations. Furthermore, we use a rectangle convolution kernel to get a larger perception range in the spatial domain.


\subsection{Convolutional sequence-to-sequence model}
\label{sec:conv-sequ-sequ}
Similar to previous works \citep{Butepage_2017_CVPR,
  Martinez_2017_CVPR}, we also use an encoder-decoder model as a
predictor to generate future motion sequences. However unlike previous
works, we adapt a convolutional model for this sequence-to-sequence
modeling task. Specifically, both the encoder and the decoder
consist of similar convolutional structure, which
computes a hidden variable based on a fixed number of inputs.
There have been several convolutional sequence-to-sequence models
\citep{bradbury2016quasi, Kalchbrenner2016NeuralMT,
  pmlr-v70-gehring17a} that have been shown to give better performance than RNN based models in machine
translation. However, these models mainly use convolution in the
temporal domain to capture correlations, while in human motion there
are also complicated spatial correlations between different body parts. 

We first
formalize the human motion prediction problem
before giving more details of our convolutional model. Assume that we are
given a series of seed human motion poses $\Xb_{1:t}=[\xb_1, \xb_2, \dots,
\xb_t]$, where each $\xb_i\in \mathds{R}^{L}$ is a parameterization of
human pose. The goal of human motion prediction is to 
generate a target
prediction $\hat \Xb_{(t+1):(t+T)}$ for the next $T$ frame poses.

We aim to capture the long-term information, such as categories of
actions, human body properties (\eg step length, step pace \etc),
environmental constraints 
\etc from the seed human motion poses. To this end, a convolutional long-term encoder is
used in our model. It maps the whole sequence $\Xb_{1:t}=[\xb_1, \xb_2, \dots, \xb_t]$ to a hidden variable
\begin{align}
  \label{eq:encoding_le}
  \zb_l^e = h_l^e(\Xb_{1:t}|\wb_l^e),
\end{align}
where $\wb_l^e$ is the parameter of the long-term encoder $h_l^e$.

Our decoder has an encoder-decoder structure, which consists of a short-term encoder and a spatial decoder. The short-term encoder 
\begin{align}
  \label{eq:encoding_se}
  \zb_s^e = h_s^e(\Xb_{t-C+1:t}|\wb_s^e),
\end{align}
where $\wb_s^e$ is the parameter. It maps a shorter sequence
$\Xb_{t-C+1:t}$, which consists of $C$ neighboring frames of the current frame, to a hidden variable. Note that our short-term encoder is a sliding window of size $C$, it only encodes the most recent $C$ frames.
Finally, the
long-term and short-term hidden variables $\zb_l^e$ and $\zb_e^e$ are
concatenated together as a input of the spatial decoder, which predicts the
next pose $\hat \xb_{t+1}$ as
\begin{align}
  \label{eq:decoder}
  \hat \xb_{t+1} = h_d([\zb_l^e, \zb_s^e]|\wb_d),
\end{align}
where $\wb_d$ is the parameter of the spatial decoder $h_d$. 
To
predict a sequence, the short-term encoder will slide one frame forward once a new frame is generated, thus the short-term encoder and decoder are applied recursively as
\begin{align}
\label{eq:shortenc-dec}
  \zb_s^e(k) &= h_s^e(\bar \Xb_{t-C+k:t+k-1}|\wb_s^e),\notag\\
  \hat \xb_{t+k}& = h_d([\zb_l^e, \zb_s^e(k)]|\wb_d),
\end{align}
where
\begin{align}
  \label{eq:bar_xb}
  \bar \Xb_{t-C+k:t+k-1}& = [\xb_{t-C+k},\xb_{t-C+k}, \ldots,
                          \xb_{t},\notag\\
 & \quad \hat \xb_{t+1}, \ldots, \hat \xb_{t+k-1}]. 
\end{align}

In our model, the long-term encoder and short-term encoder have the
similar structure, i.e. the CEM, which
includes 3 convolutional layers and 1 fully connected layers. The number of output channels for each convolutional layer is 64, 128 and 128, and the output number of the fully connected layer is 512.
As
discussed earlier, the CEM needs to capture
long-term correlations in order to improve the prediction
accuracy. Thus the stride of every convolutional layer is set to
$2$. With such convolutional layers, two elements of distance
$n$ are able to interact with each other with a path length
$\Ocal(\log (n))$, while $\Ocal(n)$ steps are required in a conventional RNN. 

Furthermore, the perception range of the CEM in the spatial domain should be large enough to capture the spatial correlations of joints from different limbs.  
Hence, we
use a rectangle $2\times 7$ convolutional kernel ($2$ along temporal
domain, and $7$ along spatial domain) to enlarge the perception domain
in the spatial domain.   
We use a simple two layer fully-connected neural
network for the spatial decoder. The first layer maps a 1024 dimensional hidden variable to 512 dimensions and uses a leaky ReLU as the activation function. The
second layer maps the hidden variable to one frame of human poses, and does not include an activation function. We also use a residual link in
our network
as suggested by previous works
\citep{Martinez_2017_CVPR,he2016deep}. This means that out decoder actually predicts the residual value rather than directly generates the next frame. Consequently, the output of our network consists of two parts: 
\begin{align}
  \label{eq:2}
   \hat \xb_{t+k}& = h_d([\zb_l^e, \zb_s^e(k)]|\wb_d) + \hat \xb_{t+k-1}.
\end{align}
$h_d$ and $w_d$ denote the decoder and its parameters. 

\paragraph{Comparison to RNN} In
recurrent neural network based 
models (\eg \citep{sutskever2014sequence,fragkiadaki2015recurrent,Martinez_2017_CVPR}), the encoded hidden variable
often serves as the initial state of the decoding RNN. Thus
during the long propagation path in RNN, the encoded information may
vanish. However, our proposed model does not have this problem
because the encoded hidden variable $\zb_l^e$
is always maintained. In recurrent neural networks, the model
captures short-term dynamical information through variation of hidden states. In
our model, the short-term dynamical information is captured by the short-term
encoder from a short sequence. By using such a structure, our model is
able to capture long-term invariant information and short-term
dynamical information, and thus resulting in better performance in both
long-term and short-term predictions.

\subsection{Optimization}
\label{sec:optimize}
During training, we use the mean squared error of the predicted poses as the loss function:
\begin{align}
  \label{eq:loss}
  &\ell_{\textrm{model}}(\hat \Xb_{(t+1):(t+T)}, \Xb_{(t+1):(t+T)})
  \notag\\
  = & \frac{1}{T}\sum_{t'=1}^{T}\|\hat  \xb_{t+t'}- \xb_{t+t'}\|_2^2.
\end{align}
Three different types of regularizing technique
are used to prevent overfitting - dropout, $\ell_2$ regularizer and adversarial
regularizer. We added a dropout layer between the
last convolutional and first fully-connected layers in our CEM module. In
our decoder, we added a dropout layer between the two fully-connected
layer. The dropout probability in both dropout layers is set to
$0.5$. 

Motivated by the generative adversarial network (GAN), we
apply an adversarial regularizer for the proposed model, which mainly improves the qualitative performance. We
train an additional discriminator to classify the generated 
and real sequences as follows
\begin{align}\label{eq:gan_dis}
  \min_{\wb_D}&-\sum_{\Xb_{1:{t+T}}}\log D(\Xb_{1:t+T}|\wb_D) \\
  &- 
  \sum_{[\Xb_{1:t}, \hat \Xb_{t+1:t+T}]}\log(1- D([\Xb_{1:t}, \hat \Xb_{t+1:t+T}]|\wb_D)).\notag 
\end{align}
The discriminator $D$ is then used to encourage the generation of realistic 
sequences.

Finally, the objective becomes
\begin{align}
  \label{eq:4}
  \min_{\wb^e_l,\wb^e_s, \wb_d}&
  \sum_{\Xb_{1:{t+T}}}\ell_{\textrm{model}}(\hat \Xb_{(t+1):(t+T)},
  \Xb_{(t+1):(t+T)})\notag\\
 & + \lambda_2[\|\wb_l^e\|_2^2 + \|\wb_s^e\|_2^2 +
   \|\wb_d\|_2^2] \\
  &- \lambda_{\mathrm{adv}} \log (D([\Xb_{1:t}, \hat \Xb_{t+1:t+T}]|\wb_D))\notag,
\end{align}
where the weights 
$\lambda_2$ and $\lambda_{\mathrm{adv}}$ are set to $0.001$ and $0.01$, respectively. 
In the optimization procedure, we used stochastic gradient
descent based optimizer to run iteratively optimizing over
\eqref{eq:4} and \eqref{eq:gan_dis}. 

\paragraph{Remarks} There are multiple choices of $\bar
\Xb_{t-C+k:t+k-1}$ in \eqref{eq:bar_xb}, which may have different effects on the training
results. In previous works, the corresponding part is often set to
ground truth, or ground truth with noise
\citep{fragkiadaki2015recurrent}. Besides setting $\bar
\Xb_{t-C+k:t+k-1}$ as \eqref{eq:bar_xb}, it can also be set to
\begin{align}
  \label{eq:5}
  &\bar \Xb_{t-C+k:t+k-1} = [\xb_{t-C+k},\xb_{t-C+k+1}, \ldots,
                          \xb_{t},\\
 & \quad \eta\hat \xb_{t+1}+(1-\eta)\xb_{t+1}, \ldots, \eta\hat \xb_{t+k-1}+(1-\eta)\xb_{t+k-1}],\notag
\end{align}
where $\eta\in[0, 1]$ is a manually specified parameter. 

Note that the window size of the short-term encoder $C$ may also affect the results. The model may not capture enough short-term information when $C$ is too small. On the other hand, it may be a waste of computation when $C$ is too large since we already have the long-term encoder. Hence, the value of $C$ should be a trade-off between accuracy and computation. The effect of different window sizes are explored in our experiments.

\section{Experiments}
\label{sec:experiments}
In this section, we apply the proposed convolutional model on several human
motion prediction tasks. The proposed method is compared with several
recent and state-of-the-art matching algorithms:
\begin{itemize}\setlength{\itemsep}{0pt}
\item The Encoder-Recurrent-Decoder (ERD) method \citep{fragkiadaki2015recurrent};
\item An three layer LSTM with linear encoder and decoder (LSTM-3LR)
  \citep{fragkiadaki2015recurrent};
\item  Stuctural Recurrent Neural Networks (SRNN)
  \citep{jain2016structural};
\item Residual Recurrent Neural Networks (RRNN)
  \citep{Martinez_2017_CVPR};
\item An three layer LSTM with an denoising auto encoder (LSTM-AE)
  \citep{ghosh2017learning}. 
\end{itemize}

Our model is implemented in tensorflow, and we used the ADAM \citep{kingma2014adam} optimizer
to optimize over our model. The batch size is set to 64 and the
learning rate is 0.0002. For more optimizing details, please refer
to Section \ref{sec:optimize}. Following the setting of previous
works\citep{fragkiadaki2015recurrent,jain2016structural}, the length of seed pose sequence is set to 50, and the length
of target sequence is set to 25. We trained RRNN
\citep{Martinez_2017_CVPR} model based on the public available
implementation\footnote{\url{https://github.com/una-dinosauria/human-motion-prediction}}. 
We quote the results from
\citep{Martinez_2017_CVPR} 
for ERD, LSTM-3LR, SRNN，, and \citep{ghosh2017learning} for LSTM-AE.

\paragraph{Action specific vs. general model} ERD \citep{fragkiadaki2015recurrent}, LSTM-3LR \citep{fragkiadaki2015recurrent} and SRNN \citep{jain2016structural} are action specific models, where they train a specific model for each action. On the other hand, the RRNN model \citep{Martinez_2017_CVPR} considers the more challenging task of training a general model for multiple actions. In our experiments, we also train a single model for multiple actions.
\begin{table*}[t]
  \centering
  \caption{\small Motion prediction error in terms of Euler angle
    error for walking, eating, smoking and discussion in the Human3.6M
    dataset for short-term of 80, 160, 320, 400, and long-term of
    1000ms (best result in bold).}
  \label{table1}
  \small
  \setlength{\tabcolsep}{3pt}
  \begin{tabularx}{0.99\textwidth}{cccccc|ccccc|ccccc|ccccc}
    \noalign{\hrule height 1pt}      
    &\multicolumn{5}{c}{Walking}\vline
    &\multicolumn{5}{c}{Eating}\vline
    &\multicolumn{5}{c}{Smoking}\vline
    &\multicolumn{5}{c}{Discussion}\\
    ms & 80 & 160 & 320 & 400 & 1000  
       & 80 & 160 & 320 & 400 & 1000  
       & 80 & 160 & 320 & 400 & 1000  
       & 80 & 160 & 320 & 400 & 1000\\
    \noalign{\hrule height 0.5pt}
    \scriptsize ERD\citep{fragkiadaki2015recurrent} 
      & 0.93 & 1.18 & 1.59 & 1.78 & N/A  
      & 1.27 & 1.45 & 1.66 & 1.80 & N/A  
      & 1.66 & 1.95 & 2.35 & 2.42 & N/A  
      & 2.27 & 2.47 & 2.68 & 2.76 & N/A\\
    \scriptsize LSTM-3LR\citep{fragkiadaki2015recurrent} 
      & 0.77 & 1.00 & 1.29 & 1.47 & N/A  
      & 0.89 & 1.09 & 1.35 & 1.46 & N/A  
      & 1.45 & 1.68 & 1.94 & 2.08 & N/A  
      & 1.88 & 2.12 & 2.25 & 2.23 & N/A\\
    \scriptsize  SRNN \citep{jain2016structural} 
      & 0.81 & 0.94 & 1.16 & 1.30 & N/A  
      & 0.97 & 1.14 & 1.35 & 1.46 & N/A  
      & 1.45 & 1.68 & 1.94 & 2.08 & N/A  
      & 1.22 & 1.49 & 1.83 & 1.93 & N/A\\
    \scriptsize   RRNN \citep{Martinez_2017_CVPR} 
      & 0.33 & 0.56 & 0.78 & 0.85 & 1.14  
      & 0.26 & 0.43 & 0.66 & 0.81 & 1.34  
   	  & 0.35 & 0.64 & 1.03 & 1.15 & 1.83 
      & 0.37 & 0.77 & 1.06 & 1.10 & \textbf{1.79}\\
    \scriptsize LSTM-AE\citep{ghosh2017learning} &1.00&
1.11&
      1.39&
            N/A & 
      1.39 &
            
    1.31 &
1.49 &
1.86 &
N/A &
2.01 & 0.92 &
1.03 &
1.15 &
N/A & 
      1.77 &
             1.11 &
1.20 &
1.38 &
N/A & 
1.73
    \\
    \scriptsize   Ours 
    	& \textbf{0.33} & \bf 0.54 & \bf 0.68 & \textbf{0.73} & \textbf{0.92}  
        & \textbf{0.22} & \textbf{0.36} & \textbf{0.58} & \textbf{0.71} & \textbf{1.24}  
        & \textbf{0.26} & \textbf{0.49} & \bf 0.96 & \textbf{0.92} & \textbf{1.62} 
        & \textbf{0.32} & \bf 0.67 & \bf 0.94 & \textbf{1.01} & 1.86\\
    \hline
  \end{tabularx}
\end{table*}

\begin{table*}
  \small
  \centering
     \caption{Motion prediction error in terms of Euler angle error
       for the rest actions in the Human3.6M dataset for short-term of
       80, 160, 320, 400, and long-term of 1000ms (best result in bold).}
  \label{table2}
  \setlength{\tabcolsep}{3pt}
  \begin{tabularx}{0.97\textwidth}{cccccc|ccccc|ccccc|ccccc}
  \noalign{\hrule height 1pt}    
    &\multicolumn{5}{c}{Directions}\vline
    &\multicolumn{5}{c}{Greeting}\vline
    &\multicolumn{5}{c}{Phoning}\vline
    &\multicolumn{5}{c}{Posing}\\
    ms & 80 & 160 & 320 & 400 & 1000  
            & 80 & 160 & 320 & 400 & 1000  
                 & 80 & 160 & 320 & 400 & 1000  
                      & 80 & 160 & 320 & 400 & 1000\\\hline
    \scriptsize{RRNN \citep{Martinez_2017_CVPR}} 
    & 0.44 & 0.70 & 0.86 & 0.97 & 1.59  
            & 0.55 & 0.90 & 1.34 & 1.51 & 2.03  
                 & 0.62 & 1.10 & 1.54 & 1.70 & 1.89  
                      & 0.40 & 0.76 & 1.37 & 1.62 & 2.56\\
    \scriptsize{Ours} 
    & \textbf{0.39} & \textbf{0.60} & \textbf{0.80} & \textbf{0.91} & \textbf{1.45}  
            & \textbf{0.51} & \textbf{0.82} & \textbf{1.21} & \textbf{1.38} & \textbf{1.72}  
                 & \textbf{0.59} & \textbf{1.13} & \textbf{1.51} & \textbf{1.65} & \textbf{1.81}  
                      & \textbf{0.29} & \textbf{0.60} & \textbf{1.12} & \textbf{1.37} & \textbf{2.65}\\

    \noalign{\hrule height 0.75pt}  
    &\multicolumn{5}{c}{Purchases}\vline 
    &\multicolumn{5}{c}{Sitting}\vline
    &\multicolumn{5}{c}{Sittingdown}\vline
    &\multicolumn{5}{c}{Takingphoto}\\
    ms & 80 & 160 & 320 & 400 & 1000  
       & 80 & 160 & 320 & 400 & 1000  
       & 80 & 160 & 320 & 400 & 1000  
       & 80 & 160 & 320 & 400 & 1000\\\hline

    \scriptsize{RRNN \citep{Martinez_2017_CVPR}} 
    	& \textbf{0.59} & \textbf{0.83} & 1.22 & 1.30 & \textbf{2.30}  
        & 0.47 & 0.80 & 1.30 & 1.53 & 2.14  
        & 0.50 & 0.96 & 1.50 & 1.72 & 2.72  
        & 0.32 & 0.63 & 0.98 & 1.12 & 1.51\\
    \scriptsize{Ours} 
    	& 0.63 & 0.91 & \textbf{1.19} & \textbf{1.29} & 2.52  
        & \textbf{0.39} & \textbf{0.61} & \textbf{1.02} & \textbf{1.18} & \textbf{1.67}  
        & \textbf{0.41} & \textbf{0.78} & \textbf{1.16} & \textbf{1.31} & \textbf{2.06}  
        & \textbf{0.23} & \textbf{0.49} & \textbf{0.88} & \textbf{1.06} & \textbf{1.40}\\
   
    \noalign{\hrule height 0.75pt}  
    &\multicolumn{5}{c}{Waiting}\vline
    &\multicolumn{5}{c}{Walkingdog}\vline
    &\multicolumn{5}{c}{Walkingtogether}\vline
    &\multicolumn{5}{c}{\textbf{Average}}\\
    ms & 80 & 160 & 320 & 400 & 1000  
            & 80 & 160 & 320 & 400 & 1000  
            & 80 & 160 & 320 & 400 & 1000 
            & 80 & 160 & 320 & 400 & 1000\\\hline

    \scriptsize{RNN \citep{Martinez_2017_CVPR}} 
    	& 0.35 & 0.68 & 1.14 & 1.34 & \textbf{2.34}  
        & \textbf{0.55} & \textbf{0.91} & \textbf{1.23} & \textbf{1.35} & \textbf{1.86}  
        & 0.29 & 0.59 & 0.86 & 0.92 & 1.42  
        & 0.43 & 0.75 & 1.12 & 1.27 & 1.90\\
    \scriptsize{Ours} 
    	& \textbf{0.30} & \textbf{0.62} & \textbf{1.09} & \textbf{1.30} & 2.50  
        & 0.59 & 1.00 & 1.32 & 1.44 & 1.92  
        & \textbf{0.27} & \textbf{0.52} & \textbf{0.71} & \textbf{0.74} & \textbf{1.28}  
        & \textbf{0.38} & \textbf{0.68} & \textbf{1.01} & \textbf{1.13} & \textbf{1.77}\\ \hline
   
  \end{tabularx} 
\end{table*}

\subsection{Dataset and Preprosessing}
\label{sec:dataset}
In the experiments, we consider two datasets: the Human 3.6M dataset \citep{h36m_pami} and the CMU Motion Capture dataset \footnote{Available at \url{http://mocap.cs.cmu.edu}}.

The Human 3.6M dataset is currently the
largest available video pose dataset, which provides accurate 3D body
joint locations recorded by a Vicon motion capture system. It is
regarded as one of the most challenging datasets because of the large
pose variations performed by different actors. There are 15 activity
scenarios in total. Each action scenario includes 12 trials lasting between 3000 to 5000 frames. The 12 trials are categorized as 6 subjects, where each subject includes 2 trials.
Each 3D pose consists of
32 joints plus a root orientation and displacement  
represented as an exponential map.  

During the experiments, each pose would subtract to the mean pose over all
trials and gets divided by the standard deviation.  We eliminate the
joint angle dimensions with constant standard deviation, which corresponds to
joints with less than three degrees of freedom. Furthermore, the global
rotation and translation are set to zero since our models are not trained with this information. 
Finally, the dimension of the
input vector is set to 54. Similar to
\citep{fragkiadaki2015recurrent,jain2016structural,Martinez_2017_CVPR}, we
treat the two sequences in subject 5 as the test set and all others as the training
set. For evaluation, we calculate the Euclidean error in terms of Euler
angle. Specifically, we measure the Euclidean
distance between our predictions and the ground truth in terms of Euler angle for each action, followed by
calculating the mean value over all sequences which are randomly
selected from the test set. 

We also apply our model to the CMU Motion Capture dataset
in order to test its generalization ability. There are five main categories in the
dataset - ``human interaction'', ``interaction with
environment'', ``locomotion'', ``physical activities \& sports'' and
``situations \& scenarios''. We choose some of the actions for our
experiments based on some criteria. Firstly, we do not use data from 
the ``human interaction'' category since multiple subjects motion
prediction is out of the scope of this paper. Secondly, action
categories which include less than six trials are excluded on the consideration
that we need enough data for each action to train our model. Lastly,
some action categories in the dataset are actually combinations of
other actions, e.g. actions in the subcategory ``playground''
consist of jump, climb and other actions which already exist in the
dataset. We do not chose these action categories to avoid
repetition. Finally, eight actions are selected for our experiments - running, walking and jumping from category ``locomotion'', basketball and soccer from category ``physical activities \& sports'',
wash windows from category ``common behaviours and expressions'',
traffic direction and basketball signals from category ``communication
gestures and signals''. We pre-process the data and evaluate the results in
the same way as we did on the Human 3.6M dataset.

\begin{figure*}[t]
  \centering
  \begin{tabular}{cc}；
    \includegraphics[width=0.48\textwidth]
    {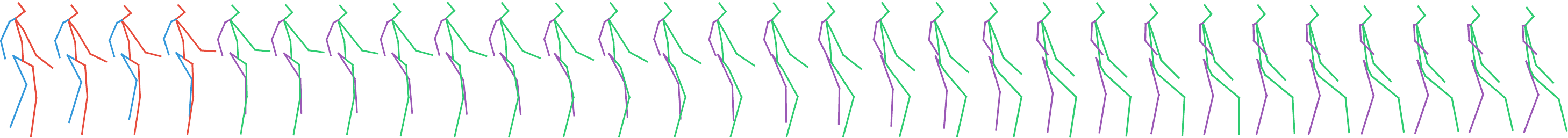}
    & \includegraphics[width=0.48\textwidth]
        {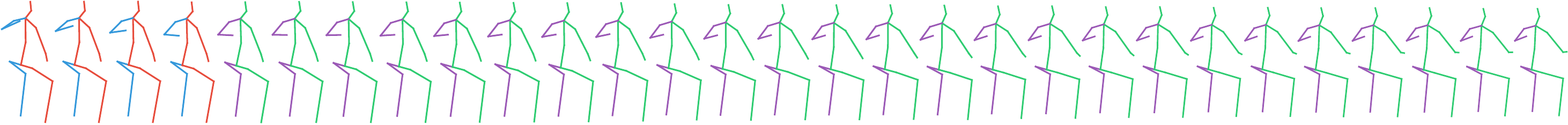}\\
    \includegraphics[width=0.48\textwidth]
    {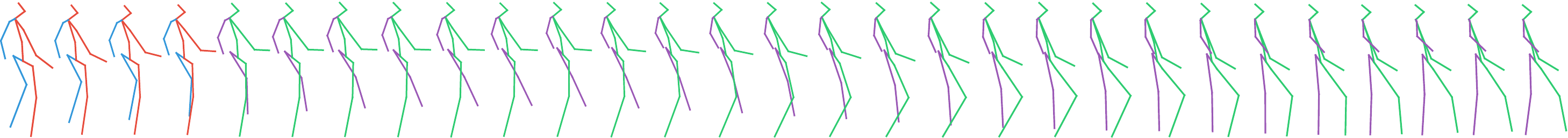}
    & \includegraphics[width=0.48\textwidth]
      {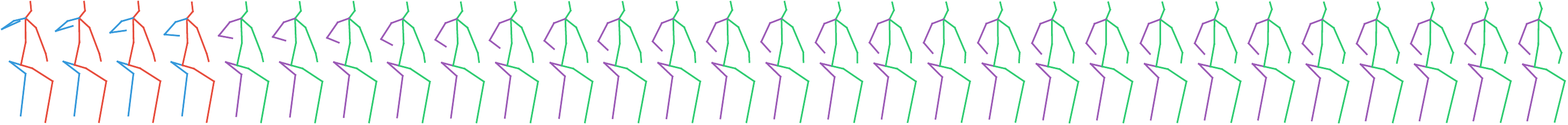}\\
    \includegraphics[width=0.48\textwidth]
    {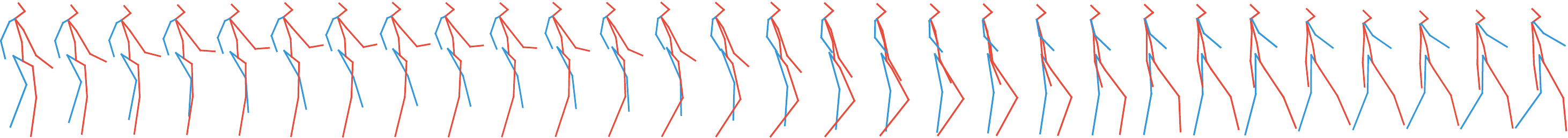}
    & \includegraphics[width=0.48\textwidth]
      {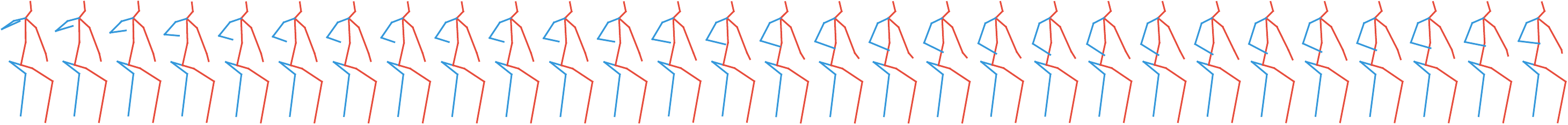}\\
    \\
     \includegraphics[width=0.48\textwidth]
        {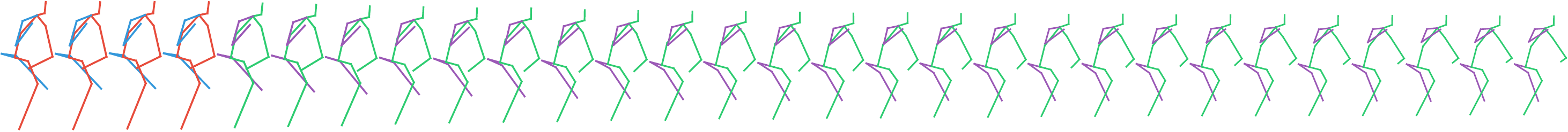}
    & \includegraphics[width=0.48\textwidth]
        {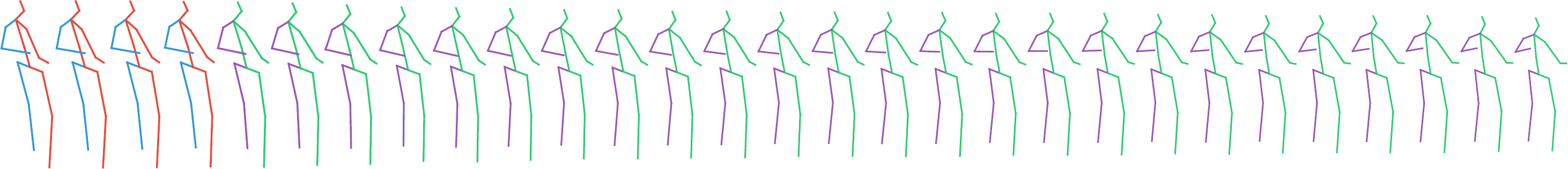}\\
     \includegraphics[width=0.48\textwidth]
    {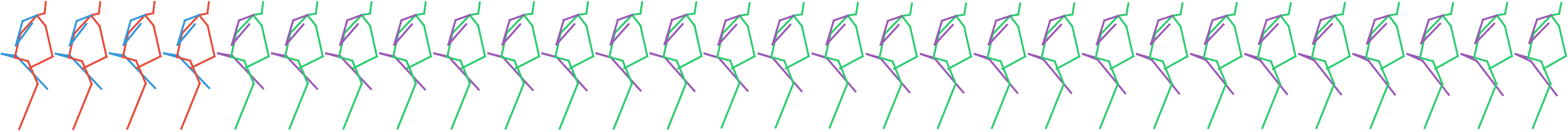}
    & \includegraphics[width=0.48\textwidth]
      {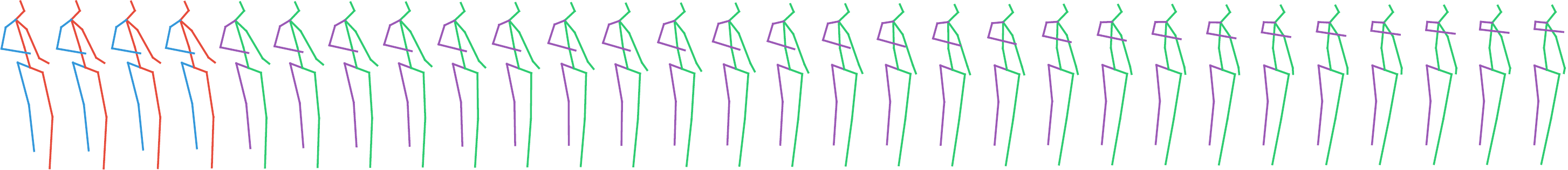}\\
    \includegraphics[width=0.48\textwidth]
    {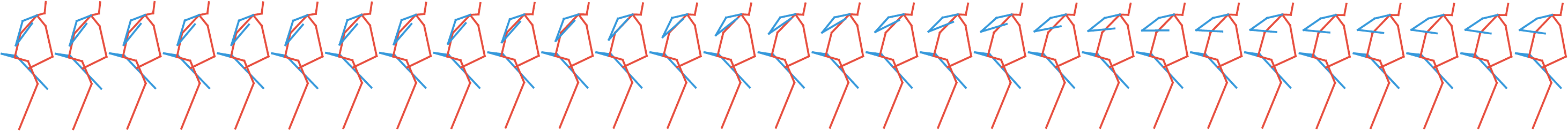}
    & \includegraphics[width=0.48\textwidth]
      {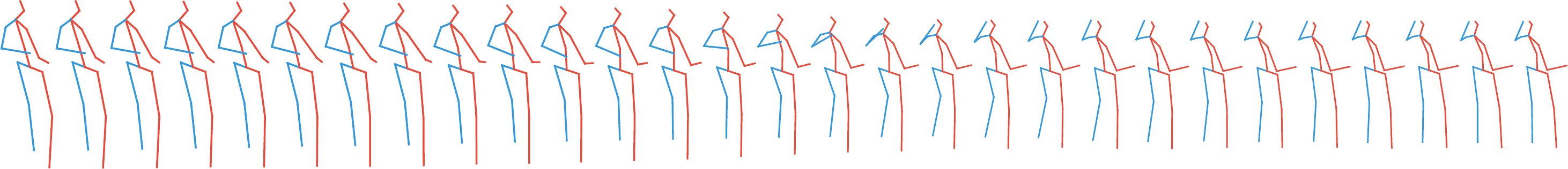}\\
  \end{tabular}
   \caption{Qualitative results on for long-term prediction based on
    the Human3.6M dataset. Starting from the left top clockwisely the
     four actions are ``walking'', ``sitting'', ``smoking'' and
     ``discussion''.  For each action, the top, middle and bottom
     sequences correspond to RRNN, our model and ground truth
     respectively. The first four frames are the last four frames of  conditional seed frames and the next ones are predicted frames. The RRNN converges to mean pose for eating and discussion, and generates a prediction which is not realistic for smoking. Our model suffers less from the mean pose problem and predicts more realistic future.}
  \label{fig:results}
\end{figure*}
\begin{figure*}[t]
  \centering
\quad\quad\quad  \includegraphics[width=0.28\textwidth]{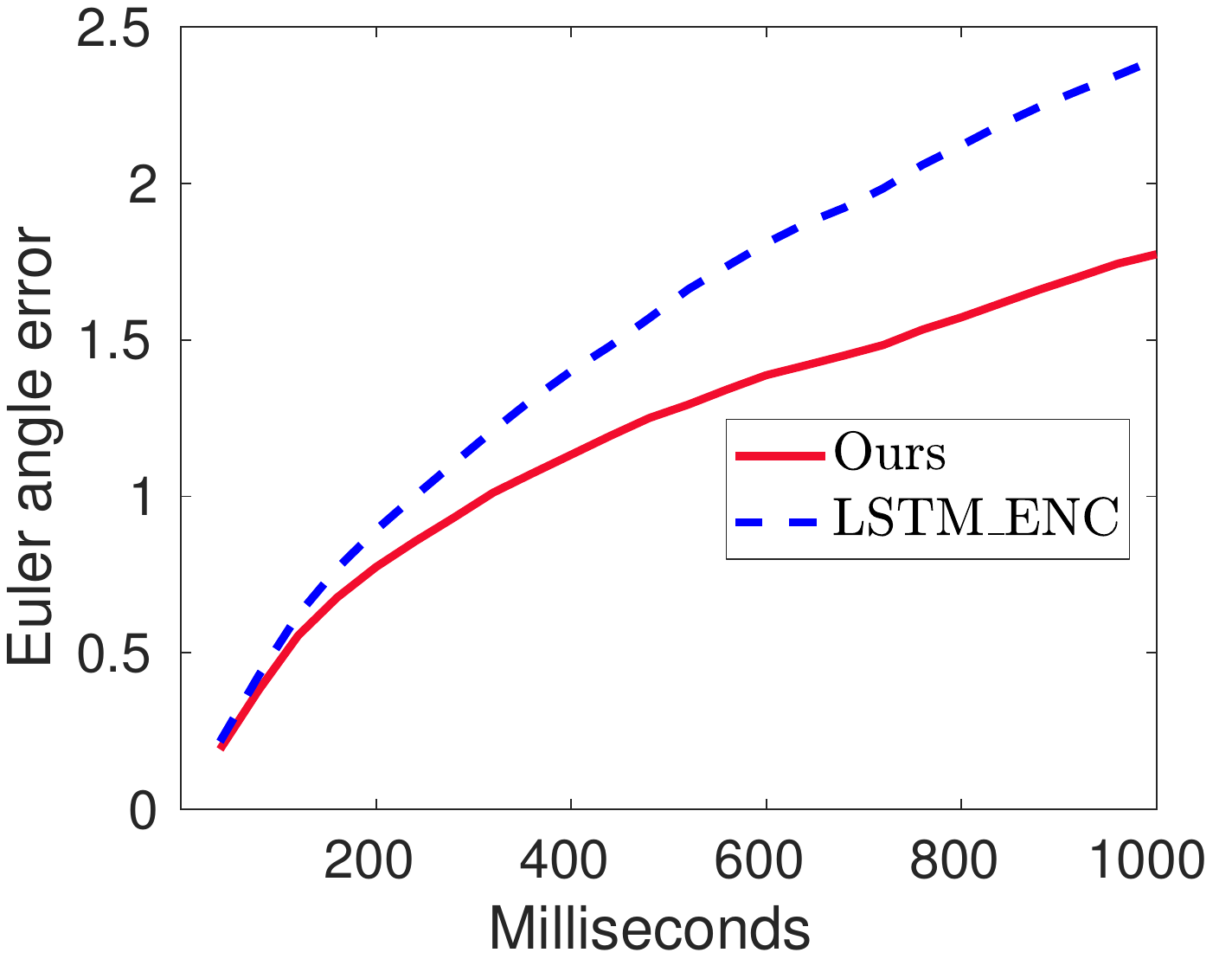} \hfill
  \includegraphics[width=0.28\textwidth]{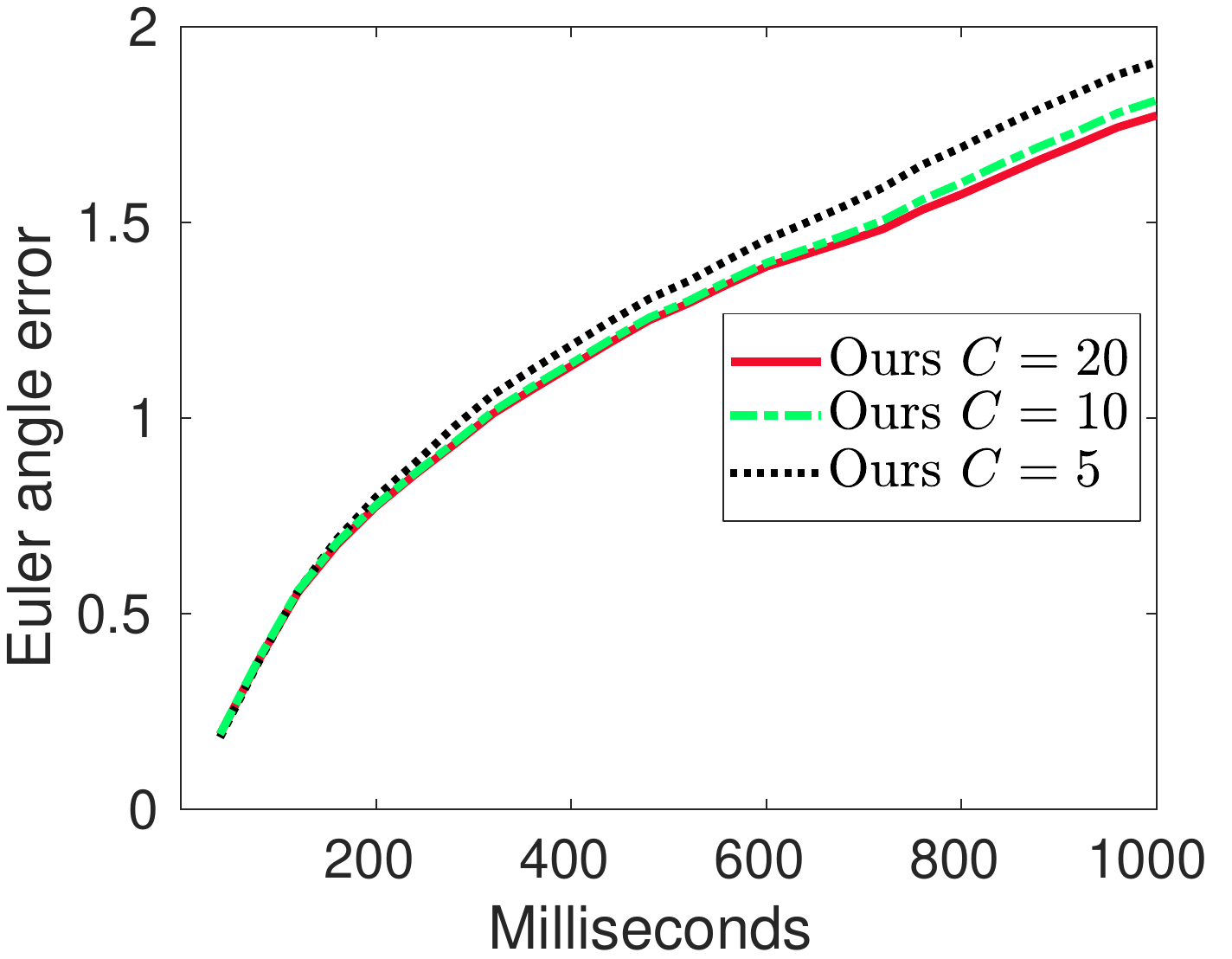}
\hfill
\includegraphics[width=0.28\textwidth]{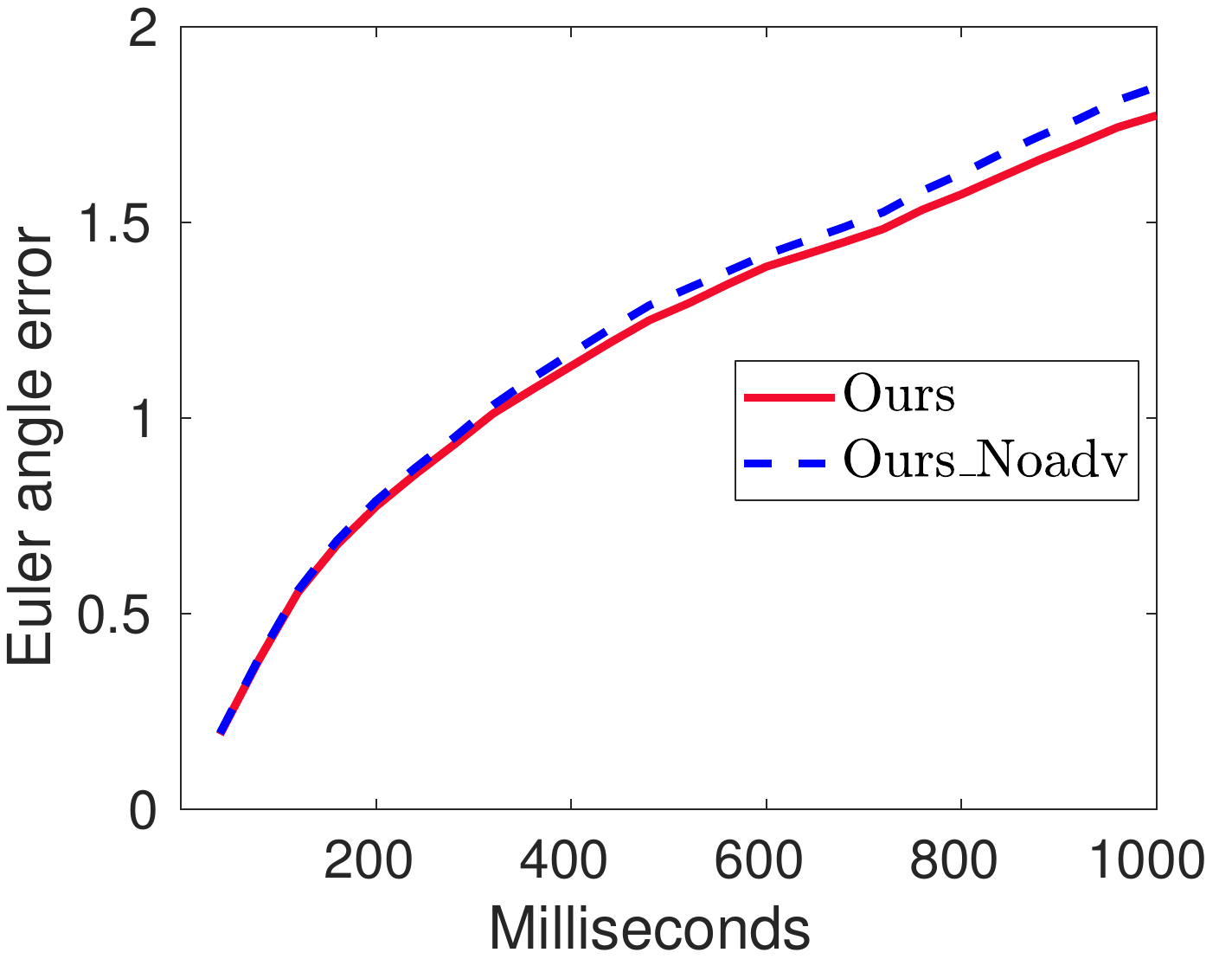}\quad\quad\quad
  \caption{\textbf{Left:} Prediction error vs time for the model. For the
    model with LSTM as encoder (LSTM\_ENC), the error accumulates much
    faster than the original model,  which means that our model performs better especially for long-term prediction.
  \textbf{Middle:} Comparison of testing error  with different window length
  $C$ for training, namely 5, 10, 20. The long-term error decreases when window size increases from 5 to 10, while the improvement is not obvious when further increase from 10 to 20.
  \textbf{Right :} Compare the testing error of our model with and without adversarial regularizer, it shows that adversarial regularizer helps to train a model with better performance in long-term.
  }
  \label{fig:eulerangle}
\end{figure*}
\begin{table*}
  \centering
  \caption{Motion prediction error in terms of Euler angle error for
    eight actions in the CMU Motion capture dataset for short-term of 80, 160, 320,
    400, and long-term of 1000ms (best results in bold).}  
  \label{table3}
    \small
  \setlength{\tabcolsep}{3pt}
  \begin{tabularx}{0.97\textwidth}{cccccc|ccccc|ccccc|ccccc}
     \noalign{\hrule height 1pt}  
    &\multicolumn{5}{c}{Basketball}\vline 
    &\multicolumn{5}{c}{Basketball Signal}\vline
    &\multicolumn{5}{c}{Directing Traffic}\vline
    &\multicolumn{5}{c}{Jumping}\\
    ms & 80 & 160 & 320 & 400 & 1000  
       & 80 & 160 & 320 & 400 & 1000  
       & 80 & 160 & 320 & 400 & 1000  
       & 80 & 160 & 320 & 400 & 1000\\\hline

    \scriptsize{RRNN \citep{Martinez_2017_CVPR}} 
    	& 0.50 & 0.80 & 1.27 & 1.45 & \bf1.78  
        & 0.41 & 0.76 & 1.32 & 1.54 & 2.15  
        & 0.33 & 0.59 & 0.93 & 1.10 & 2.05  
        & 0.56 & 0.88 & 1.77 & 2.02 & 2.40\\
    \scriptsize{Ours} 
    	& \textbf{0.37} &\bf 0.62 & \bf 1.07 & \bf 1.18 & 1.95  
        & \textbf{0.32} & \bf0.59 & \textbf{1.04} & \textbf{1.24} & \bf1.96  
        & \bf 0.25 &\bf  0.56 &\bf  0.89 &\bf  1.00 &\bf  2.04  
        & \bf 0.39 & \bf 0.60 &\bf  1.36 &\bf 1.56 & \textbf{2.01}\\

    \noalign{\hrule height 0.75pt}   
    &\multicolumn{5}{c}{Running}\vline
    &\multicolumn{5}{c}{Soccer}\vline
    &\multicolumn{5}{c}{Walking}\vline
    &\multicolumn{5}{c}{Washwindow}\\
    ms & 80 & 160 & 320 & 400 & 1000  
       & 80 & 160 & 320 & 400 & 1000  
       & 80 & 160 & 320 & 400 & 1000  
       & 80 & 160 & 320 & 400 & 1000\\\hline

    \scriptsize{RRNN \citep{Martinez_2017_CVPR}} 
    	& 0.33 & 0.50 & 0.66 & 0.75 & 1.00  
        & 0.29 & 0.51 & 0.88 & 0.99 & 1.72  
        &\bf 0.35 & 0.47 & 0.60 & 0.65 & 0.88  
        & 0.30 & 0.46 & 0.72 & \bf 0.91 &\bf 1.36\\
    \scriptsize{Ours} 
    	& \bf \bf 0.28 & \bf 0.41 & \bf 0.52 & \bf 0.57 & \bf 0.67  
        & \bf 0.26 & \bf 0.44 &\bf  0.75 & \bf 0.87 &\bf  1.56  
        &\bf 0.35 & \bf 0.44 & \bf 0.45 & \bf  0.50 & \bf  0.78  
        & \bf 0.30 & \bf 0.47 & \bf 0.80 &  1.01 & 1.39\\
	\hline
  \end{tabularx}
\end{table*}

\begin{figure*}[t]
  \centering
   \includegraphics[width=0.9\textwidth]
        {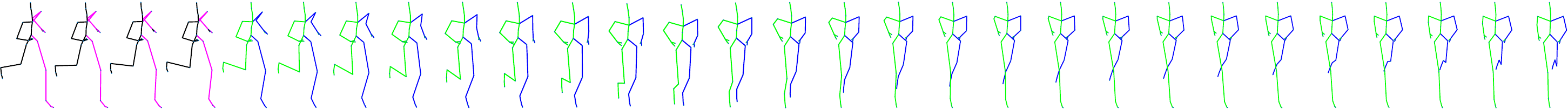}\\
     \includegraphics[width=0.9\textwidth]
      {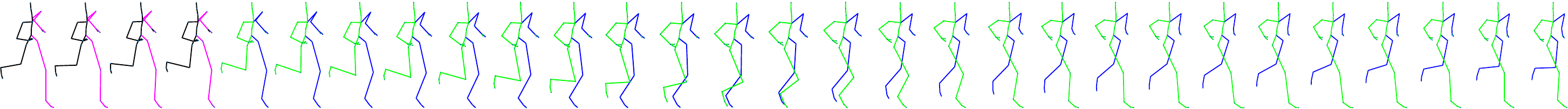}\\
     \includegraphics[width=0.9\textwidth]
      {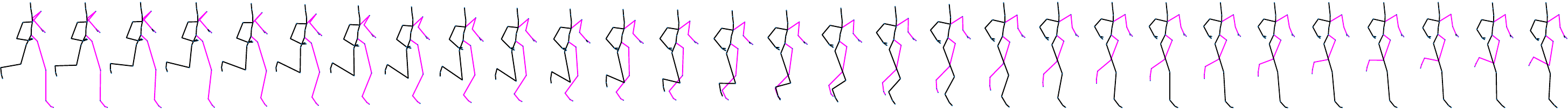}\\
\vspace{0.4em}

        \includegraphics[width=0.9\textwidth]
    {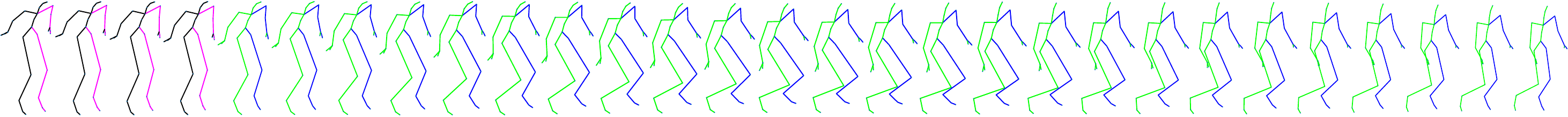}
       \includegraphics[width=0.9\textwidth]
        {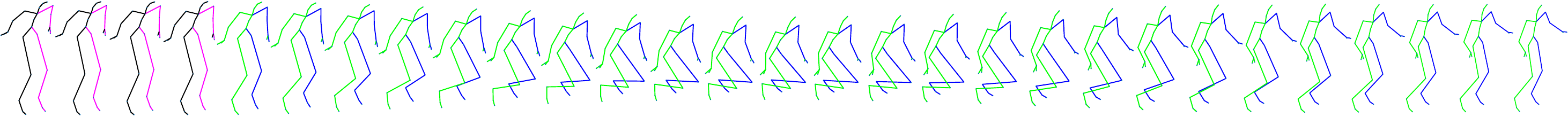}
     \includegraphics[width=0.9\textwidth]
         {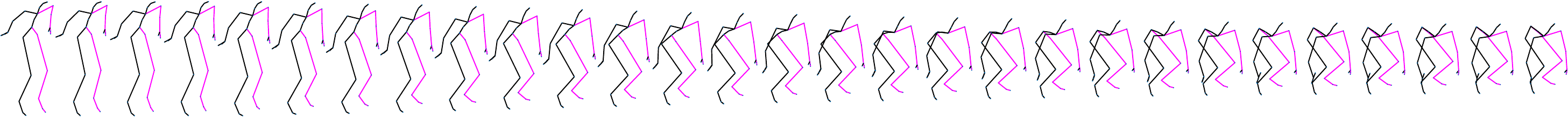}


  \caption{Qualitative result on the CMU Motion Capture
    dataset. \textbf{Top:} ``Running'' action; \textbf{Bottom: }
    ``Jumping'' action.  For each action, the top, middle are
    the results of RRNN and our model, and the ground truth is given in
     the bottom.
    The first four frames are the last four frames of the conditional seed frames and the next are the predicted ones.}
  \label{fig:results_on_cmu}\vspace{-1em}
\end{figure*}
\subsection{Evaluation on Human3.6M and CMU Datasets}
We first report our results on all actions in the Human 3.6M dataset for
both short-term prediction of 80 ms, 160 ms, 320 ms, 400 ms and
long-term prediction of 1000 ms. Among the 15 actions in the dataset,
the four actions ``walking'', ``eating'', ``smoking'' and
``discussion'' are commonly used in comparison of action specific
human motion prediction methods. Thus we compare the accuracy of our
method against four action specific methods ERD
\citep{fragkiadaki2015recurrent}, LSTM-3LR
\citep{fragkiadaki2015recurrent}, Structural RNN (SRNN)
\citep{jain2016structural} and LSTM-AE \citep{ghosh2017learning}, as well as one general motion prediction method RRNN\citep{Martinez_2017_CVPR} in Table \ref{table1}. From the results, we can see that our method outperforms the others in most cases. We also provide the qualitative comparison results with the state-of-the-art RRNN method
in Figure \ref{fig:results}. Both RRNN and our model achieve good result on ``walking'' because of its periodic property, which makes the action easier to model. But for other aperiodic classes like ``eating'', ``smoking'' and ``discussion'', RRNN quickly converges to a mean pose -- the predicted figure could not put its hands down in ``eating'' and raise its hand up in ``discussion''. 

Rather than maintaining a gesture in which one leg should be put on the other one
in the action ``smoking", RRNN generates an implausible motion in real life that would cause the subject to go off balance. 
This further shows that it is very important to take the correlation between different body parts into consideration so that the predicted pose is more realistic. In comparison, our model predicts plausible motions for both ``eating'' and ``smoking''. 
Furthermore, it is observed in the highly aperiodic action ``discussion'' that our model 
can still predict the correct motion trend, i.e. raising the hands while talking, even though this motion is not exactly the same as the ground truth.
We compare our algorithm with the general human prediction model RRNN \citep{Martinez_2017_CVPR} for the other 11 actions. The quantitative comparison results are provided in Table \ref{table2}, which suggest that our algorithm outperforms RRNN in most cases. Additionally, 
our method outperforms RRNN on the average in both long and short-term predictions. The out-performance of our method becomes more significant for longer term predictions
.

\begin{table}[t]
\small
\centering
\caption{Average testing error on the CMU Motion Capture dataset in terms of Euler angle error.}\label{tabe:avg_err_cmu}
\begin{tabularx}{0.40\textwidth}{cccccc}
    \noalign{\hrule height 1pt}  
    ms & 80 & 160 & 320 & 400 & 1000 \\
                \noalign{\hrule height 0.75pt}  
    \scriptsize{RRNN \citep{Martinez_2017_CVPR}} 
    	& 0.38 & 0.62 & 1.02 & 1.17 & 1.67 \\
    \scriptsize{Ours} 
    	& \bf 0.31 & \bf 0.52 & \bf 0.86 &\bf  0.99 &\bf 1.55\\
            \noalign{\hrule height 0.5pt}  
\end{tabularx}
\end{table}
We only consider the more challenging
task of training a general motion prediction model for all actions 
using the CMU Motion Caption dataset. Hence, we only show comparison results with the state-of-the-art RRNN method.
For a fair comparison, both our model and RRNN are
trained using the same settings on the Human3.6M dataset. The testing error of each action is given in Table \ref{table3} and the average testing error is given in Table \ref{tabe:avg_err_cmu}. In the quantitative comparison, our method outperforms the RRNN method in several challenging actions such as jumping and running. 
The qualitative comparisons of running and jumping are also shown in Figure \ref{fig:results_on_cmu}. In the qualitative comparisons, we can see that RRNN converges to mean pose for both running and jumping. On the other hand, 
our prediction for running is very close to the realistic one. For jumping, our model also predicts the correct motion trend, i.e. squatting followed by jumping, and the main error comes from 
the duration of squatting.

\subsection{Ablation Study}
\noindent \textbf{The role of long-term encoder} The long-term encoder in our model is used to capture long-term dependencies. We verify its effectiveness 
by removing it from our model. The results in 
Table \ref{Tab:AblationStudyLTE} suggest that the average error gets larger without the long-term encoder, especially for long-term prediction of 1000 ms. 

\vspace{0.2cm}
\noindent \textbf{Rectangular kernel over spatial axis} We use a rectangular kernel over the spatial axis ($2 \times 7$ kernel) in our CEM in order to better capture the dependencies between different body parts. We also verify the effectiveness by comparing it with square kernel ($4 \times 4$ kernel) and rectangular kernel over the temporal axis ($7 \times 2$ kernel). The results 
in Table \ref{Tab:AblationStudyRK} 
indicate that the $2 \times 7$ kernel  
is the best choice.
\begin{table}[t]\small
	\centering
	\caption{\small Our model w/wo long-term Encoder on Human3.6M.}
	\begin{tabular}{cccccc}
		\hline
		ms & 80 & 160 & 320 & 400 & 1000 \\  
		\hline      
		\scriptsize{With long-term Encoder} & \textbf{0.38} & \textbf{0.68} & \textbf{1.01} & \textbf{1.13} & \textbf{1.77} \\ 
		\scriptsize{Without long-term Encoder} & 0.41 & 0.72 & 1.05 & 1.17 & 1.88\\ 
		\hline
	\end{tabular}\label{Tab:AblationStudyLTE}
\end{table}

\vspace{0.2cm}
\noindent \textbf{Adversarial regularizer} We use an adversarial regularizer in our model to help generate more plausible motion. In order to explore the role of the adversarial regularizer, we compare the performance of our model with and without the regularizer. The result in Figure \ref{fig:eulerangle} suggests that the adversarial regularizer helps to improve the performance of our model even though marginally. Moreover, since the adversarial regularizer is only used during training, it does not add complexity to our model during inference.

\vspace{0.2cm}
\noindent \textbf{Different window size $C$} In our decoder, different window sizes $C$ result in different perception range. Intuitively, enlarging the window size may enlarge the perception range and results in better performance, but also requires more computation resources. We thus train three different models with window size $C=5$， $10$ and $20$. In the right plot of Figure \ref{fig:eulerangle}, we show average testing error over all 15 actions. The result suggests that there is not much improvement when the window size is larger than 10. Consequently, we set $C=20$ for our model in view of the trade-off between accuracy and computational accuracy.


\begin{table}[t]\small
	\centering
	\caption{\small Comparison of different kernels on Human3.6M.}
	\begin{tabular}{cccccc}
		\hline
		ms & 80 & 160 & 320 & 400 & 1000 \\  
		\hline         
		\scriptsize{$4\times 4$ kernel} & 0.41 & 0.72 & 1.05 & 1.16 & 1.80\\ 
		\scriptsize{$7\times 2$ kernel} & 0.40 & 0.71 & 1.05 & 1.17 & 1.79\\ 
		\scriptsize{$2\times 7$ kernel} & \textbf{0.38} & \textbf{0.68} & \textbf{1.01} & \textbf{1.13} & \textbf{1.77}\\ 
		\hline 
	\end{tabular}\label{Tab:AblationStudyRK} 
\end{table} 

\section{Conclusion}
\label{sec:disc-future-works}
In this work, we proposed a convolutional sequence-to-sequence model for human motion prediction. We adopted two types of convolutional encoders in our model, namely the long-term encoder and short-term encoder, so that both distant and nearby temporal motion information can be used for future prediction. 
We demonstrated that our model performs better than existing state-of-the-art RNN based models, especially for long-term prediction tasks. Moreover, we show that our model can generate better predictions for complex actions
due
to the use of 
hierarchical convolutional structure 
for modeling 
complicated spatial-temporal correlations.

\section*{Acknowledgements}
This work was partially supported by the Singapore MOE Tier 1 grant R-252-000-637-112 and the National University of Singapore AcRF grant R-252-000-639-114. 

{\small
  \bibliographystyle{ieee}
  \bibliography{ref}
}
\appendix
\end{document}